\documentclass[preprint,12pt]{elsarticle}

\usepackage{amssymb}
\usepackage{amsmath}
\usepackage[linesnumbered,ruled,vlined]{algorithm2e}
\usepackage{algpseudocode}

\usepackage{epsfig}
\usepackage{color,soul}
\setstcolor{yellow}
\usepackage{multirow}
\usepackage[table,xcdraw]{xcolor}

\usepackage[table,xcdraw]{xcolor}
\usepackage[]{changes}
\usepackage{nomencl}
\renewcommand{\nomgroup}[1]{%
  \item[\textbf{%
    \ifthenelse{\equal{#1}{M}}{Parameters}{}%
    \ifthenelse{\equal{#1}{N}}{Variables}{}%
    \ifthenelse{\equal{#1}{I}}{Sets and Indices}
    }]%
}
\makenomenclature

\usepackage[acronym,nomain,nonumberlist]{glossaries}
\makeglossaries
\newacronym{dr}{DR}{Demand Response}
\newacronym{iso}{ISO}{Independent System Operator}
\newacronym{sm}{SM}{Smart Meter}
\newacronym{ems}{EMS}{Energy Management System}
\newacronym{rl}{RL}{Reinforcement Learning}
\newacronym{db}{DB}{Demand Bidding}
\newacronym{mcp}{MCP}{Market  Clearing  Price}
\newacronym{dbe}{DBE}{Demand Bidding Event}
\newacronym{cbl}{CBL}{Customer Baseline Load}
\newacronym{mdp}{MDP}{Markov  Decision  Process}
\newacronym{ddpg}{DDPG}{Deep Deterministic Policy Gradient}
\newacronym{dqn}{DQN}{Deep Q Network}
\newacronym{dso}{DSO}{Distribution System Operator}
\newacronym{tso}{TSO}{Transmission System Operator}

\journal{Applied Energy}
\begin{document}

\begin{frontmatter}



\title{Data-Driven Online Interactive Bidding Strategy for 
Demand Response }


\author[A]{Kuan-Cheng Lee}
\author[A]{Hong-Tzer Yang}
\author[B,C]{Wenjun Tang*}

\address[A]{Department of Electrical Engineering, National Cheng Kung University, Tainan 70101}
\address[B]{School of Artificial Intelligence, Shenzhen Polytechnic, Shenzhen 518055}
\address[C]{Smart Grid \& Renewable Energy Lab, Tsinghua Berkeley Shenzhen Institute, Shenzhen 518055\protect\\
e-mail: monikatang@sz.tsinghua.edu.cn.}


\begin{abstract}
Demand response (DR), as one of the important energy resources in the future’s grid, provides the services of peak shaving, enhancing the efficiency of renewable energy utilization with a short response period, and low cost. Various categories of DR are established, e.g. automated DR, incentive DR, emergency DR, and demand bidding. However, with the practical issue of the unawareness of residential and commercial consumers' utility models, the researches about demand bidding aggregator involved in the electricity market are just at the beginning stage. For this issue, the bidding price and bidding quantity are two required decision variables while considering the uncertainties due to the market and participants. In this paper, we determine the bidding and purchasing strategy simultaneously employing the smart meter data and functions. A two-agent deep deterministic policy gradient method is developed to optimize the decisions through learning historical bidding experiences. The online learning further utilizes the daily newest bidding experience attained to ensure trend tracing and self-adaptation. Two environment simulators are adopted for testifying the robustness of the model. The results prove that when facing diverse situations the proposed model can earn the optimal profit via off/online learning the bidding rules and robustly making the proper bid.

\end{abstract}

\begin{graphicalabstract}

\includegraphics[width=0.9\columnwidth]{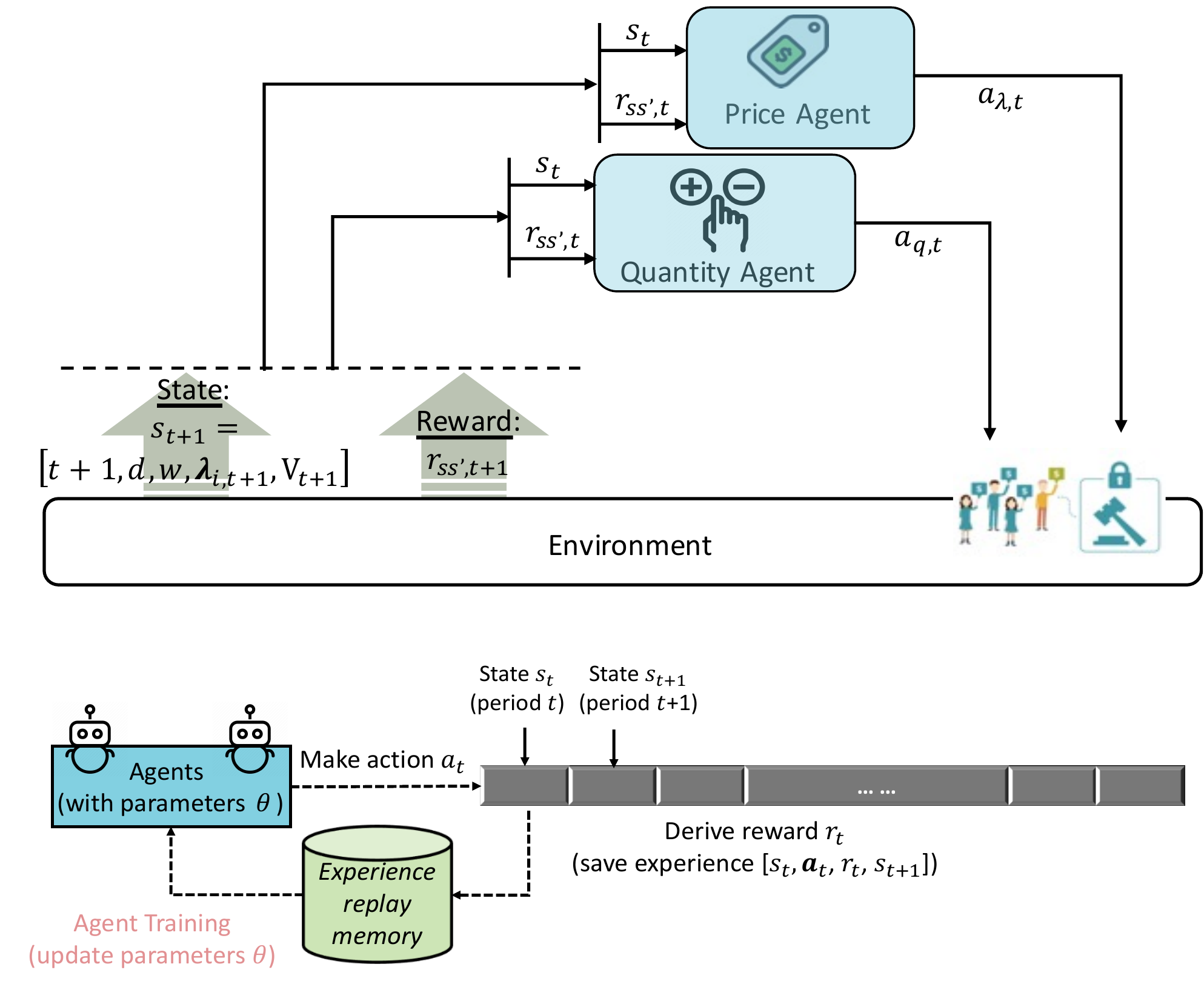}
\end{graphicalabstract}



\end{frontmatter}
\nomenclature[N]{$\lambda_{bid,t}$}{The bidding price of aggregator}%
\nomenclature[N]{$q_{bid,t}$}{The bidding quantity of aggregator}%
\nomenclature[M]{$t^s$}{The starting time of Demand Bidding Event}%
\nomenclature[M]{$t^e$}{The ending time of Demand Bidding Event}%
\nomenclature[I]{$\Lambda_i$}{The customer $i$’s participation plan}%
\nomenclature[M]{$\lambda_{i,t}$}{The price of the customer $i$ submitting to curtail at time $t$}%
\nomenclature[M]{$q_{act,t}$}{The actual total load sheddingat time $t$}%
\nomenclature[M]{$CBL_{i,t}$}{The $i$'s Customer Baseline Load at time $t$}%
\nomenclature[M]{$p_{i,t}$}{The actual consumption for customer $i$ at time $t$}%
\nomenclature[N]{$\xi_t$}{The execution rate at time $t$}%
\nomenclature[N]{$\xi_t$}{The execution rate at time $t$}%
\nomenclature[M]{$\lambda_{M,t}$}{Market Clearing Price at time $t$}%
\nomenclature[N]{$P$}{The profit during one DB event}%
\nomenclature[M]{$\alpha_t$}{Incentive ratio}%
\nomenclature[N]{$s_t$}{State variable}%
\nomenclature[N]{$\boldsymbol{a}_t$}{Action variable}%
\nomenclature[N]{$r$}{Reward}%
\nomenclature[N]{$\pi_\theta(\boldsymbol{a}_t|s_t)$}{Stochastic policy}%
\nomenclature[N]{$R_t$}{Cumulative discounted reward}%
\nomenclature[N]{$Q^\pi (s_t,a_t)$}{Action-value function}%


\glsaddall
\printglossary[type=\acronymtype,title=Acronyms]

\printnomenclature

\section{Introduction}
The serious environmental challenges are caused by the increasing massive consumption of fossil fuels. Meanwhile, energy shortage becomes a worldwide global issue. Take Taiwan as an example, the reserve rate at peak hour recently drops on the verge of restriction on electricity use and brings dramatic risk to the power system \citep{lu2018evaluation}. To solve the crisis issue, the development of renewable and sustainable energy sources is encouraged. 
Demand response (DR) is also adopted by many independent system operators (ISOs) with the role of flexible demand resources to improve economic efficiency and system reliability \citep{strasser2014review}.


Nowadays, most ongoing DR programs have industrial customers as the target participants but neglecting a large amount of available DR potential from residential and commercial customers, especially for the demand bidding framework\citep{8681637}. Meanwhile, the research discussing the DR program on residential and commercial customers is always set prior hypothesis, e.g., the awareness of consumers' utility function\citep{iria2018optimal} or the load is directly controllable\citep{utama2021demand,dong2021strategic}, which makes it not practical to the realistic application.   Specifically, the residential and commercial sectors accounted for about 40\% of total U.S. energy consumption in 2018 \citep{EIA2018}.  On the other hand, smart meters (SMs) and energy management system (EMS) have been vastly deployed in different phases around residential and commercial customers recently. All the above calls for the research on the DR program targeting residential and commercial loads with good practicability.

Different from the industrial customer who individually has higher average consumption and can be enrolled in DR programs directly, the residential and commercial customers need to be aggregated jointly participating in the program to have a finger in the pie. 
Generally, we name the aggregated entity as aggregator. The aggregator acts as a broker between DR resources and the system operator and a manager who aims to maximize the DR profit.

For all kinds of DR programs, demand bidding is the most complicated type, which, to our best knowledge, still lacks comprehensive problem formulation, especially for commercial and residential consumers. In this paper, we stand in the view of the aggregator considering two fundamental issues. One is the unawareness of the time-varying and consumer-varying comfort requirement of consumers while another is the unawareness of the prices in the market. Specifically, we aim to help the profit-oriented aggregator to determine its purchase strategy to the customers and its bidding strategy to the market through the data collected by Smart Meters. The contributions of our work can be concluded as:
\begin{itemize}

\item \textit{\textbf{Model-Free Strategy with Win-Win solution:}} {Employing a two-agent deep deterministic policy gradient method, our Reinforcement Learning (RL)-based strategy model generates a solution benefiting for both aggregator and consumers without the awareness of consumers' utility model;}
\item \textit{\textbf{Uncertainties Consideration:}} The uncertainties, no matter coming from the price in the market or from the comfort requirement among consumers, are simultaneously measured in the decision-making model. Through learning from the growing amount of experience, the model obtains the strategies approaching the profit maximized;
\item \textit{\textbf{Self-Adjusting and Learning Ability:}} With the state-of-the-art RL method and structure of dynamic learning, the proposed two-agent model can adapt to the implicit market trend and the consumers' comfort preference variation by the online training structure;
\item \textit{\textbf{Robustness in Variation:}} We design an environment simulator to evaluate the robustness of our model. The results exhibit the adaptivity and the profit-pursuing ability of the model under various uncertainties. Moreover, both the online and offline models perform stable decision-making abilities especially after more and more experience is involved.
\end{itemize}

The remainder of this paper is organized as follows. We review the related works and research in Section \ref{LR}. We detail the framework and the proposed methodology in Section \ref{FW} and \ref{PF}.
We discuss the simulation results in Section \ref{Sim} and conclude this paper in Section \ref{con}

\section{Literature Review}
\label{LR}
Generally DR management is considered as a demand-side management problem
\citep{bahrami2017online,parizy2018low,lu2012evaluation}
, especially when formulating the problem in price-based DR. The loads of the consumer are distinctively divided into 2 categories: elastic and inelastic type. These two types can further  be detailed to the elastic appliances with a memory-less property, elastic appliances with a full memory property, elastic appliances with a partial memory property, inelastic appliances with an interruptible operation, and inelastic appliances with an uninterruptible operation \citep{7579628}.
Based on the classification, the works in \citep{bahrami2017online} develop the methods to schedule the elastic and interruptible loads, i.e., appliances, to gain the benefit from DR. Most of these assume the load can be directly controlled and thus are designed either for peak load reduction \citep{parizy2018low} or regulation services \citep{lu2012evaluation}. Indirect load control methods are typically extensions of direct load control methods that include additional control components to convert incentive signals, e.g., prices, to comfort signals, e.g., thermostat offsets \citep{korkas2016occupancy}
. Simultaneously, the comfort of users participating in DR is counted in the scheduling, represented by minimum waiting time of appliances or thermal comfort set in the constraints or the utility of consuming in the objective function. However, how to quantify comfort is still a big issue, as even to the user itself, the understanding of self-consuming requirements lacks a comprehensive knowledge for varying times and dates, especially for the residential and commercial consumer. 

Therefore, when standing in the view of the aggregator side, the problem becomes more complicated, let alone when the aggregator joining in DB program.  To win the bids and make the profit in the DB, the uncertainties both from the market clearing price (MCP) and customers’ consumption behavior are the key factors to be considered. Some Research discusses how aggregator determines the purchase strategy to allocate the proper quantity in the DB market. The authors in \citep{lyao2018} propose a purchase strategy to find the combination of purchase status and then minimize the total bidding cost aiming to respond to the offered load curtailment. Acting as a price-taker, the aggregator in \citep{jia2019purchase} constructs a two-stage nested bilevel model for the optimal bidding strategy in day-ahead and balancing markets. However, these works assume the curtailment quantity is provided by the seller, e.g., consumer, (the same issue we mentioned before) just following the signal of incentive or curtailment. In other words, customers are required to submit their curtailment quantity and no bidding strategy is considered in the study.  P. Li et al. raise an online learning model to estimate the cost function of consumers by sending varying price signals \citep{li2017distributed}. Even though still lacking the consideration of the price uncertainty aggregator received from the utility, this work inspires us with the idea of formulating the consumer response to price through a model-free methodology.

Some works discuss the methodology to deal with the price uncertainty when bidding under the electricity market mechanism, where robust optimization \citep{tang2019optimal}, stochastic programming \citep{fazlalipour2019risk}, chance-constrained optimization \citep{zhao2018strategic}, and conditional value at risk (CVaR) \citep{alashery2019second} schemes are mostly used. A bidding strategy for the DR aggregator is proposed by \citep{7216732}. It utilizes stochastic programming to simulate different scenarios regarding price uncertainty, and then employs robust optimization with forecast values and bounded price intervas to manage uncertainty and seek solutions under different robustness. Concerning a microgrid operator who manages the intermittent distributed generation, storage, dispatchable DR, and price responsive loads, robust optimization is also utilized to manage uncertainties associated with the renewable generation intermittence, the MCP, and customers’ behavior to maximize profit or minimize operating cost \citep{tang2019optimal,liu2015bidding,rahimiyan2015strategic}. Other works use risk-constrained optimization for aggregators to deal with uncertainties by setting their acceptable risk when aggregator bids in the energy pool market \citep{asensio2015risk,xu2015risk,nguyen2014risk}. Nevertheless, the decision-makers, e.g. aggregator and system operator, in these studies are all treated as the price-taker, namely, they are fully aware of the price it can gain. Moreover, the stochastic optimization is done offline using historical data which is incompetent to handle continuous real-time bidding situations effectively. 

Therefore, a technical framework is required to implement effective DB programs from the aggregator's point of view. The framework must be both automated and able to sense and minimize user discomfort while bidding and maximizing the profit in the market-based program. Meanwhile, the correlation between the purchase and bidding strategy must be seriously counted in the framework. As a model-free learning algorithm, Reinforcement Learning (RL) is an agent-based AI algorithm where the agents learn the optimal set of actions, i.e., the optimal policy, through the interaction with the environment \citep{henderson2018deep}. The literature \citep{vazquez2019reinforcement} collects the articles focused on RL implemented DR strategy for both utility companies and consumers in detail. The RL models, e.g., on-policy and off-policy learning, have been mostly utilized in the energy systems to evaluate the human factor, i.e., comfort, desire to consume less energy, and to pay a lower energy bill \citep{wang2020deep,lu2019demand}. The work of \citep{lu2019incentive} also supports that the RL techniques can be used to maximize user comfort or minimize the energy consumption while adapted and applied to DR just by adding a few additional states, e.g., price. Moreover, RL is applied to the bidding strategy problem to decide the bidding curve to the electricity market as well \citep{rahimiyan2008supplier}. 
However, the factors affecting consumer's energy consumption and price elasticity in the previous research are still based on a great prior hypothesis. This motivates us to develop a systematic framework to generate the purchasing and bidding strategies for the aggregator in DB program simultaneously and in a model-free way.

\section{Framework of The Proposed Method and Demand Bidding Program}
\label{FW}
DB, as a market-based program, is the most complicated type of DR. For the aggregator, to participate in the DB, it is responsible to submit the bidding order, including the bidding price and quantity while considering the uncertainties that come from both market and consumers. By collecting the consumption data and price signal from consumers' sides (through SM) and the reserve and price information from the market side, our proposed scheme achieves the bidding decision.  We introduce the framework of our study and the rules of DB program we followed to construct our problem in this section. These are the foundation of our following research.

\subsection{Framework of The Proposed Bidding Strategy}

\begin{figure}[ht]
    \centering
    \includegraphics[width=1\columnwidth]{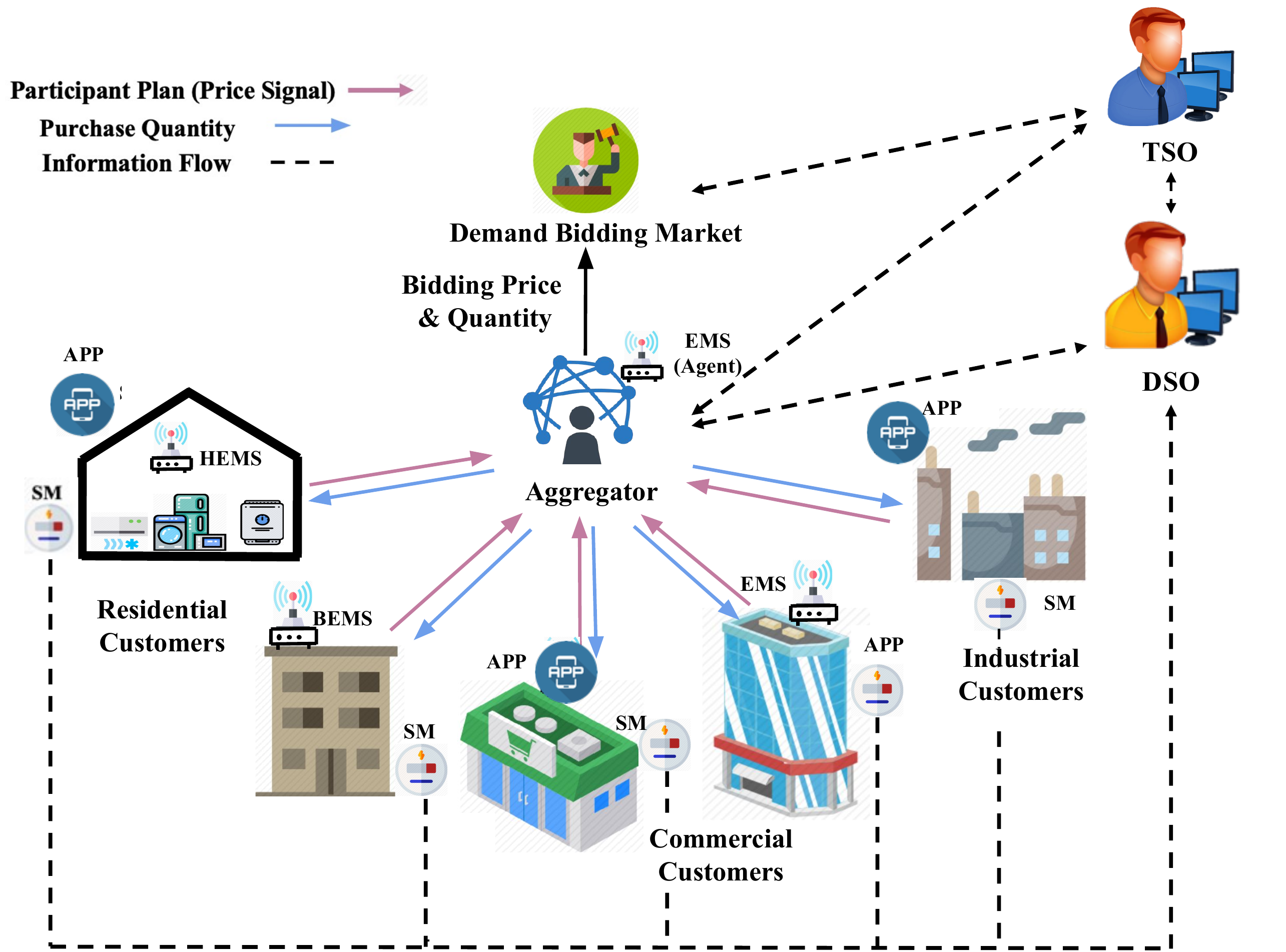}
    \caption{The Framework of The Research}
    \label{fig:overall}
    \vspace{-0.5cm}
\end{figure}

We consider residential households, commercial buildings, and small industrial facilities as the consumers who are willing to participate in the DB program through the aggregator. The overall framework is illustrated in Fig. \ref{fig:overall}. The SM is installed in each consumer and equipped with real-time electricity consumption monitoring and recording function. Bidirectional communication among the electricity consumers, the aggregator, Distribution System Operator (DSO), and Transmission System Operator (TSO) is also functioned to provide remote metering and signal reception. DSO collects the end users' consumption information in real-time and is informed of the bidding quantity of the aggregator after the DB market is cleared. Meanwhile, TSO continuously receives the information from the markets, aggregators, and DSOs which supports the maintenance of the system. Besides, each participant in our study is assumed to possess EMS, e.g. household EMS (HEMS) and building EMS (BEMS), which are all abbreviated to EMS in the following description. In this paper, the communication function of EMS enables the users to send bidding information to the aggregator and receive the execution signal of DR. It should be noted that this work concentrates the substratum strategy with the consumers, e.g., purchase strategy, and the superstratum strategy in the DB market, e.g., bidding strategy. The scheduling strategy of participants' EMS during the DR event will not be discussed in the paper. Specifically, the purchase policy of the aggregator is determined based on the bids submitted by customers to quantify the comfort level and DR potential. And the uncertainties contributed by the market clearing price are taken as the main issue when making a bidding strategy. For the situation that the consumer does have the EMS installed, an app developed for recording the meter ID-correlated participation plan can also achieve the information communication between consumer and aggregator which also works for our framework. The bidding and purchase strategy are interacted and mutually reinforcing, which is thus considered in parallel in our following model formulation. 

\subsection{DB Program}
\label{program}
\begin{figure}[ht]
    \centering
    \includegraphics[width=1\columnwidth]{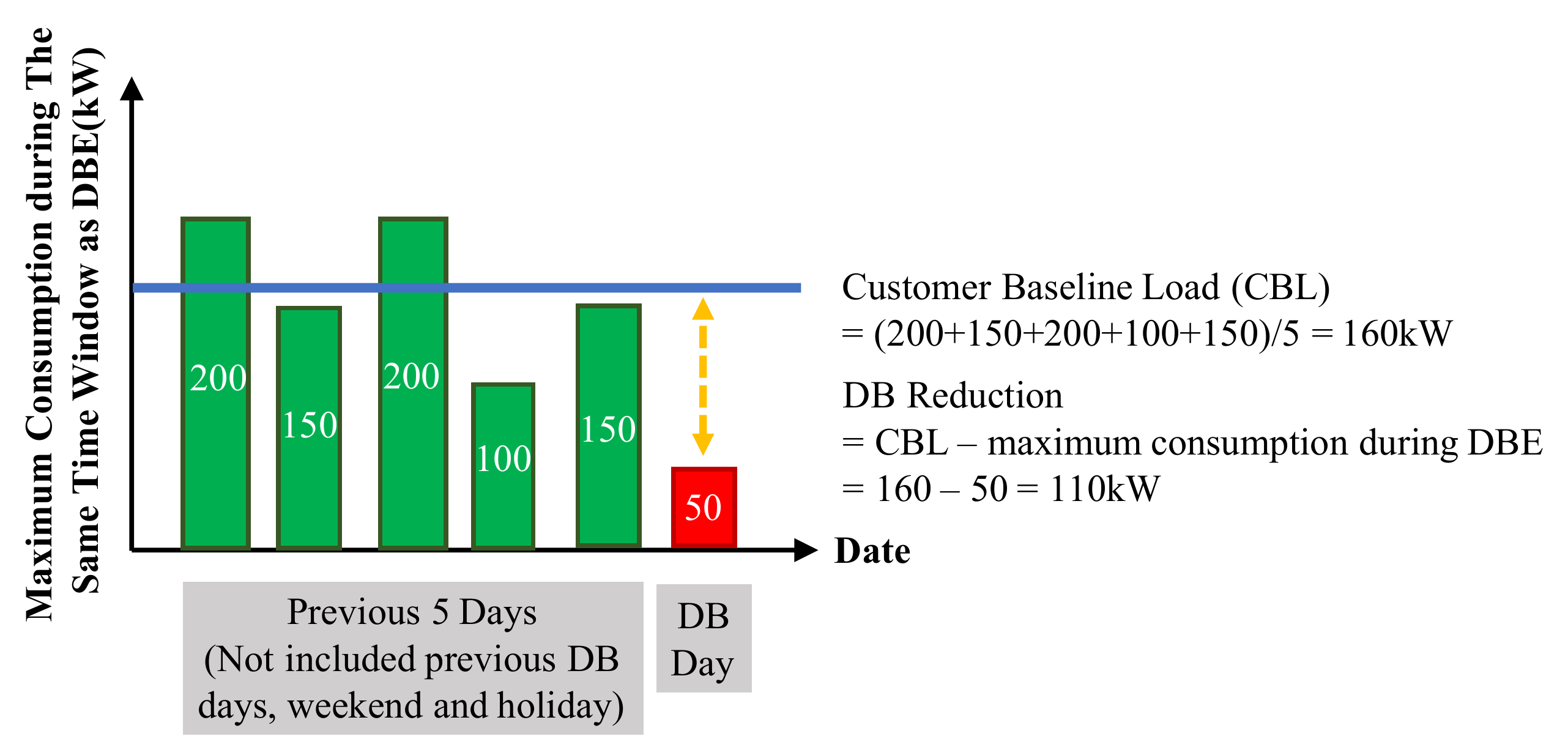}
    \caption{Example for the calculation of CBL and DB reduction}
    \label{fig:CBL}
    \vspace{-0.5cm}
\end{figure}

As mentioned in \citep{yan2018review}, dynamic price signals
can motivate consumers to adjust their electricity consumption accordingly and be combined with automation in end-use systems to have the potential to deliver more benefits to operators and consumers. We therefore modify the DB program proposed by Taipower Company \citep{website2013}, where only the high-voltage consumer 
with the contract capacity higher than 100W is allowed to participate through aggregator. The modification makes all the end-use consumers allowed to participate while the rules except the admittance threshold are kept. We have the rules detailed as below:

\begin{enumerate}[1.]
    \item The DB market is independent of other electricity markets. The DB event (DBE) is called ahead of the day when the system operator estimates an energy or system flexibility shortage.
    After the call, the participants submit the bidding price and quantity.
    The market operator clears the market by 18:00 and announces the MCP;
    \item Aggregators who bid lower than the MCP win the bidding and should practically curtail their aggregated loads and get the 
   actual total load shedding as close to the bidding quantity as possible during the DBE intervals;  
    \item The Customer Baseline Load (CBL) of each consumer is determined by the average of the maximum demand during the same time window(s) as DBE in the previous five days (with DBE days, off-peak days, and weekends excluded);
    \item The consumer's DB reduction is evaluated distinctly as the difference between the maximum actual consumption and CBL during DBE. We take an example in Figure \ref{fig:CBL} to clarify the relationship between CBL and DB reduction. The amount of the reduction collected from the consumers under an aggregator represents the actual total load shedding of the aggregator;
\item The overall incentive for each participant is calculated based on the pay-as-bid mechanism and execution rate of its bidding curtailment quantity in this DR event;
    \item The program is a risk-free program, viz. the participant gains the rewards as the requested amount is met; otherwise, no penalty is given. 
    \end{enumerate}

The rules are all formulated and the details will be given in Section \ref{PF}. It should be noted the formulation can be adapted according to the other realistic DB application, e.g., modifying the penalty in the objective function according to the rules of risky DB program. 

\section{Problem Formulation and Proposed Strategies}
\label{PF}
\subsection{System Model}

The aggregator acts as a participant in the DB market. Namely, the aggregator needs to obey the rule of DB described in Section \ref{program}. 
The bidding steps can be described as: 

The market operator firstly notifies the starting and ending time, e.g., $t^s$ and $t^e$, of the DBE before the market open. The consumers are then notified of $t^s$ and $t^e$ by the aggregator. Afterward, each consumer $i$ submits its participation plan $\Lambda_i$ to aggregator which contains the information on which periods they want to participate and the price for curtailment. The aggregator is required to submit the bid including price $\lambda_{bid,t}$ and quantity $q_{bid,t}$ for the period starting from time $t^s$ and ending at time $t^e$ in one event, where $t\in[t^s,t^{s+\triangle t},\dots,t^{s+n\triangle t},\dots,t^{s+(N-1)\triangle t}]$ and $t^s+N\triangle t=t^e$. $\triangle t$ represents the time slot and the number of time slots included in one event is counted with $n=[0,\dots,N-1]$. $\triangle t$ is a contracted value between aggregator and user, which assumed to be 15 minutes in our study and can be adjusted to meet the requirements of other cases. $t^s$ and $t^e$ are broadcasted by the market operator in advance. After winning the bid, the aggregator needs to commit the promised curtailment quantity.
The customer $i$’s participation plan $\Lambda_i$ is defined as follows. 

\begin{equation}
\small
    \Lambda_i=[\lambda_{i,t^s} ,\lambda_{i,t^{s+\triangle t}},\dots,\lambda_{i,t^{s+n\triangle t}},\dots,\lambda_{i,t^{s+(N-1)\triangle t}}]
\end{equation}

$\lambda_{i,t}$ stands for the price of the customer $i$ submitting to curtail at time $t$ while $\lambda_{i,t}=0$ means the customer $i$ has no willing to curtail at this time.

As described in the Fig. \ref{fig:overall}, the problem mainly deals with three uncertainty issues. One comes from the consumers' curtailment, one from the market price, while another from the bidding settlement. We thus detail the formulation of the problem into three concepts. 

\subsubsection{The Uncertainty Contributed by The Customers}
In the real application, most customers have no sense about their actual electricity usage, let alone their possible DR quantity. Based on the assumption that the report price from customers reflects the comfort requirement, in the framework of our method, the aggregated customers do not need to submit their DR quantity but the price $\lambda_{i,t}$. Namely, the responsibility of the DR quantity prediction is shifted from the consumer side to the aggregator side, which fits the realistic situation. Therefore, a model for predicting the total aggregated DR quantity, i.e., the bidding quantity $q_{bid,t}$, is introduced accordingly. The accuracy of the prediction depends on how well the aggregator understands the price elasticities $\epsilon_{i,t}$ of its customers, which describes how sensitive the demand changes in response to their order price $\lambda_{i,t}$ \citep{song2017price}. 

Moreover, the prediction accuracy of $q_{bid,t}$ influences the profit of the aggregator. In other words, we target to minimize the difference between prediction DB quantity $q_{bid,t}$ and the actual total load shedding $q_{act,t}$. $q_{act,t}$ is the summation of the reduction from individual $CBL_{i,t}$ to the actual consumption $p_{i,t}$ for customer $i$ as
\begin{equation}
    q_{act,t}=\sum_{i=1}^{I}q_{act,t}^i=\sum_{i=1}^{I}{max({0,CBL_{i,t}-p_{i,t}})}.
\end{equation}
The execution rate $\xi_t$ at time $t$ can then be calculated as
\begin{equation}
    \xi_t=\frac{q_{bid,t}}{q_{act,t}}.
\end{equation}
Generally, we aim to make the $\xi_t$ determined by our model-generating $q_{bid,t}$ approaching 1, i.e.,

\begin{equation}
    \lim{\xi_t}=1.
    \label{tar1}
\end{equation}

\subsubsection{The Uncertainty Contributed by The Market}
The Genco bidding problems mostly consider the factors affecting the MCP to optimize the bidding decisions \citep{li2011modeling}. The decision of bidding price $\lambda_{bid,t}$ is affected by the uncertainty of MCP $\lambda_{M,t}$. In other words, to win the bid, $\lambda_{bid,t}$ must be lower than or equal to $\lambda_{M,t}$. Meanwhile, the $\lambda_{bid,t}$ needs to approach $\lambda_{M,t}$ to earn as much profit as possible, i.e.,
\begin{equation}
    \lim{\lambda_{bid,t}}=\lambda_{M,t}. 
    \label{tar2}
\end{equation}
Conclusively, we target to decide the bidding price $\lambda_{bid,t}$ which can achieve tracking $\lambda_{M,t}$ and then maximize the profit for the aggregator.

Considering the system information generally open on the ISO website \citep{caiso}, we make the forecasting system reserve $V_t$ as the input of our decision model to optimize the bidding strategy. The reason for considering the reserve rate is because it is a key factor to $\lambda_{M,t}$. Especially when the reserve rate of the system is low, namely, the system is on the edge of restrictions of electricity use, the MCP will rise.

To consider the uncertainties from consumer side and market side, the aggregator collects the information of the customers' participation plan $\Lambda_i$ and system's forecasting reserve $V_t$. Its decision model jointly generates the bidding strategy, e.g., $\lambda_{bid,t}$, and purchasing strategy, e.g., $q_{bid,t}$ to achieve the targets, e.g., (\ref{tar1}) and (\ref{tar2}). The decision is made with the boundaries, i.e., $\lambda_{bid,t} \in [\lambda_{bid,t}^{min},\lambda_{bid,t}^{max}]$ and $q_{bid,t}\geq 0$, where $\lambda_{bid,t}=\lambda_{bid,t}^{min}=0$ represents a non-participant state.

\subsubsection{The Bidding Settlement}
Once the bidding price $\lambda_{bid,t}$ for time $t$ is determined, the aggregator’s purchase strategy is settled down. The customers, whose plan price $\lambda_{i,t}$ is lower than the bidding price $\lambda_{bid,t}$ are informed to win the bid. This strategy ensures the cost of aggregator paying to customers is lower than or equal to the profit gained from the demand bidding market. The auxiliary variable $x_{i,t}$ declaring whether customer $i$ wins the bid for time $t$ is defined as 
\begin{equation}
 x_{i,t}=\left\{
\begin{aligned}
&1, \lambda_{i,t} \leq \lambda_{bid,t},\\
&0, \lambda_{i,t} > \lambda_{bid,t},
\end{aligned}
\right.   
\end{equation}
where $ x_{i,t}=1$ means a win bid for $i$.
According to the incentive rule, the profit during one DB event for the aggregator can be calculated as the difference of the profit owning from the DB market and the cost paying to the consumers' curtailment, i.e.,
\begin{equation}
\small
    P= \\
    \sum_{t=t^s}^{t^s+N\Delta t}{(\alpha_t \times \lambda_{bid,t} \times q_{act,t}-\sum_{i=1}^I x_{i,t}\times \lambda_{i,t}\times q_{act,t}^i)\times \Delta t}, 
    \label{profit}
\end{equation}
where $\alpha_t$ denotes as the incentive ratio, which is correlated with the execution rate $\xi_t$. Specifically, in the DB program in Taiwan, the incentive ratio $\alpha_t$ is determined as
\begin{equation}
    \alpha_t=\left\{
\begin{aligned}
    &1.1, &0.8\leq \xi_t \leq 1.2, \\
    &1.05, &0.6\leq \xi_t < 0.8 \ or \ 1.2 < \xi_t \leq 1.5, \\
    &1, &0\leq \xi_t< 0.6 \ or \ 1.5 < \xi_t.
\end{aligned}
\right.
\label{intratio}
\end{equation}
We utilize the incentive ratio determined by Equation (\ref{intratio}) in the following studies. Moreover, by adjusting the incentive ratio rule according to different DB program, the calculation of $P$ can be made and further used in the proposed model. 

Considering our target of maximizing the profit $P$ in (\ref{profit}), based on the uncertainties and settlement framework introduced in this subsection, we employ a two-agent Deep  Deterministic  Policy  Gradient (DDPG) reinforcement learning method to construct our decision model and obtain the bidding and purchase strategies. 

\subsection{Markov Decision Process (MDP)-Based Two-Agent Learning Mechanism}
In this subsection, we demonstrate the proposed two-agent RL structure for the aggregator to establish the bidding and purchase policy. State, action, and reward compose the basic structure for RL, which allows the agent-based aggregator to achieve online learning-by-doing and gradually optimizing its policy. As estimates in Fig. {\ref{fig:MDP}}, the two agents in our proposed model operate in parallel while sharing the information from the environment states, utilize the same reward function, and finally obtain the policies at the same time.  We name the two agents as price agent and quantity agent separately. The former determines the bidding price $\lambda_{bid,t}$ and the later determines the bidding quantity $q_{bid,t}$. These two decision variables are involved in the model as the action $\boldsymbol{a}_t$, which generated based on the state $s_t$ and the reward $r_t$. The following definitions clarify how to embed our problem into RL formulation:
\begin{figure}[ht]
    \centering
    \includegraphics[width=0.9\columnwidth]{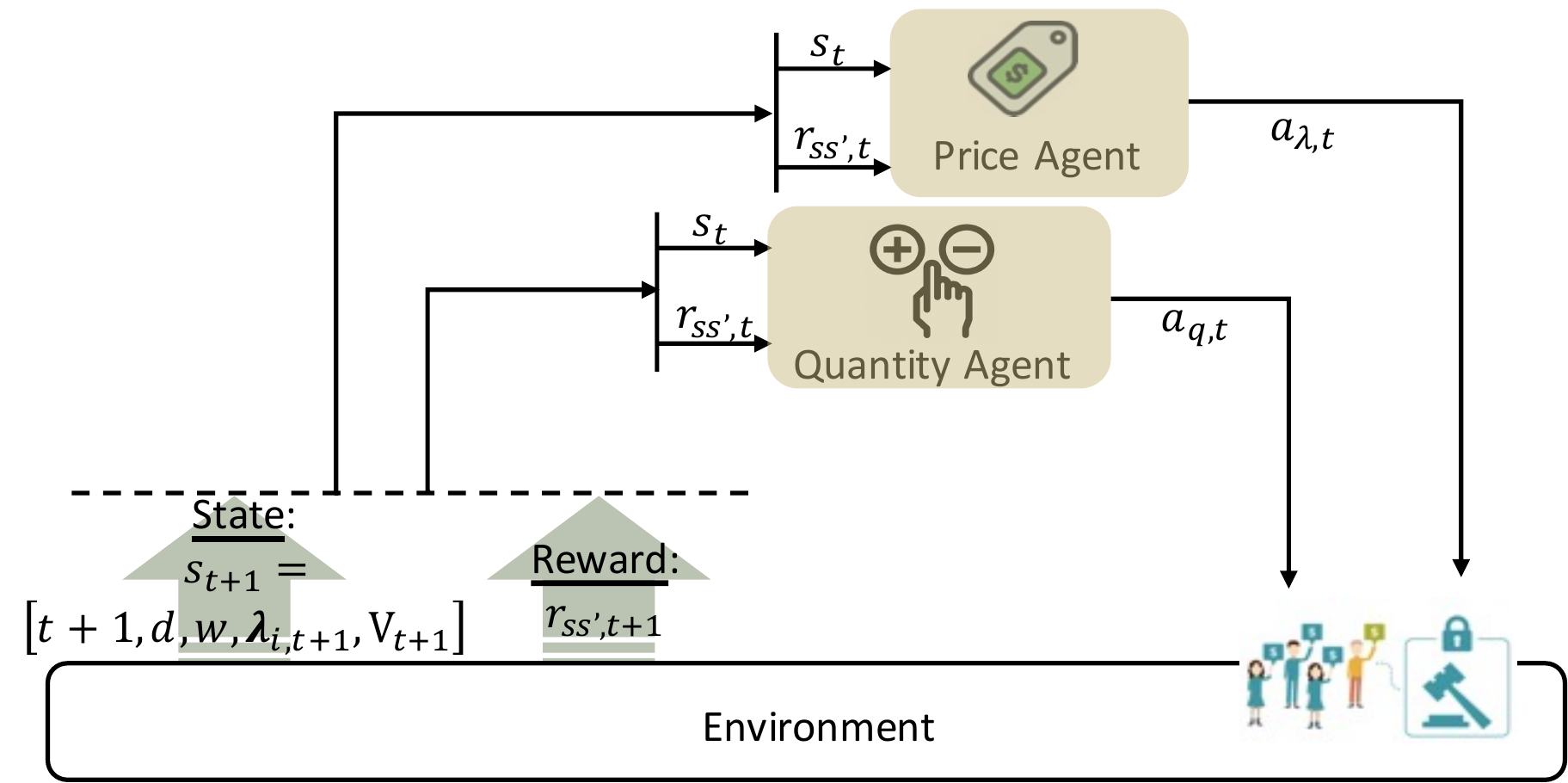}
    \caption{MDP-Based Two-Agent Learning Mechanism}
    \label{fig:MDP}
        \vspace{-0.5cm}
\end{figure}
\begin{itemize}
    \item \textit{State variable $s_t$:} The two agents share the state variable, including the information the model receiving and updating noted by time $t$. The information consists of system reserve forecast $V_t$, consumers' participation plan $\{\boldsymbol{\lambda}_{i,t}, i=1,\dots,I.\}$, and the corresponding time-related stamps, e.g., time $t$, date $d$, and weekday/weekend category $w$. For each time, the state can be described as
    \begin{equation}
        s_t=[t, d, w, \boldsymbol{\lambda}_{i,t}, V_t].
    \end{equation}
    \item \textit{Action variable $\boldsymbol{a}_t$:} The action variables composed of bidding price $a_{\lambda,t}$ and curtailment quantity $a_{q,t}$ are distinctly determined by price agent and quantity agent. The actions are generated by the policies of the two agents which equally represent the bidding price and quantity and are determined within the market constraints, i.e.,
    \begin{equation}        \lambda_{bid,t}=a_{\lambda,t},
        \ s.t., \, \lambda_{min} \leq a_{\lambda,t} \leq \lambda_{max},
    \end{equation}
     \begin{equation}
        q_{bid,t}=a_{q,t},
        \ s.t., \,  a_{q,t} \geq 0
    \end{equation}
    \item{Reward $r_t$:} The reward $r_t$ is defined referring to Equation (\ref{profit}). It is utilized and calculated jointly by the two agents as well. The reward is gained after the two actions, e.g., $a_{\lambda,t}$ and $a_{q,t}$, made while the state transfers from $s$ to $s'$, which is denoted as $\mu(a_{q,t},a_{\lambda,t},s, s')$. The expected total reward along the training episode evaluates the policy, which thus helps to improve the bidding and purchasing strategies.   
\end{itemize}

MDP provides a framework
that is widely used for modeling agent-environment interactions. An MDP can be represented by a tuple $(\boldsymbol{s},{\boldsymbol{a}},{r})$, where $\boldsymbol{s}$ and $\boldsymbol{a}$ are the two sets of the states and possible actions. We utilize the DDPG method to construct the model.

\subsection{DDPG Method}

\begin{figure}[ht]
    \centering
    \includegraphics[width=1\columnwidth]{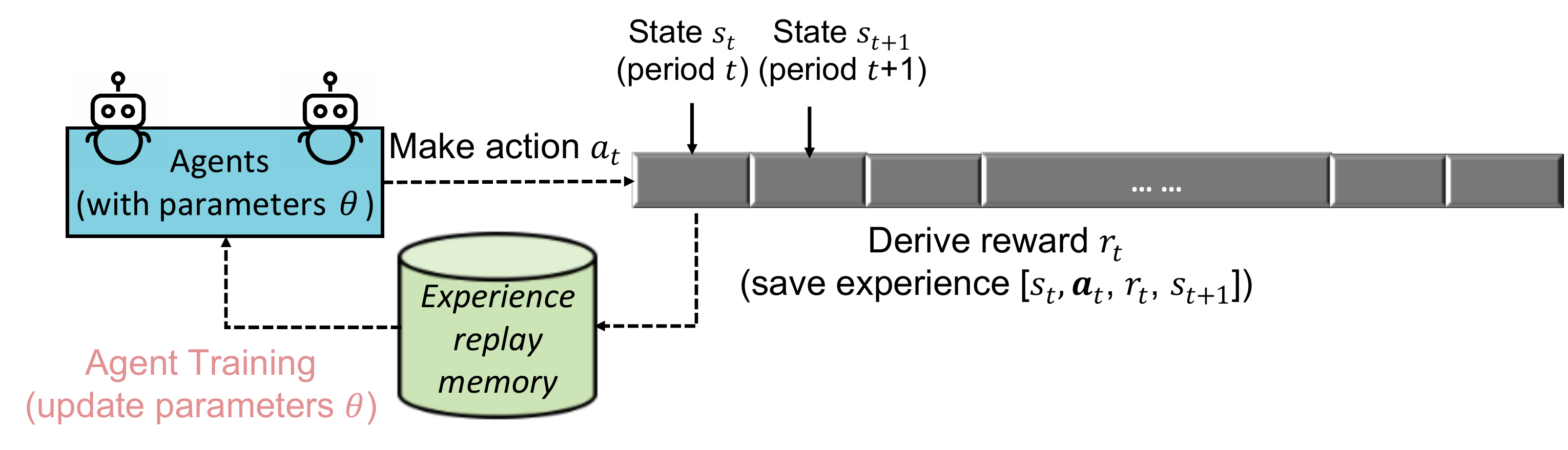}
    \caption{Online Learning Process for Agents}
    \label{fig:online}
        \vspace{-0.5cm}
\end{figure}

In this paper, DDPG learning method \citep{lillicrap2015continuous} is used to solve the agent-based aggregator bidding model. As an off-policy RL model, DDPG not only combines the advantages of DQN, policy gradient, and actor-critic but also adopts the deterministic policy to accelerate the converge time. Fig. \ref{fig:online} shows how agents interact with the environment to attain experience and then instantly updates their parameters to optimize the actions. The agents constantly store the newest experience to achieve online learning so that it can adapt to the implicit market trend in real-time. With the same learning structure, offline and online learning models are constructed sequentially. The offline learning uses a large amount of historical data to train the model iteratively. And online learning utilizes the trained model to make the strategies and also real-timely learns the up-to-date experience during the DR event.

\textit{1) Preliminary}

We construct the agent with a neural network structure to act and learn in a stochastic environment by sequentially making actions over time steps under the MDP structure. We model the DB problem as MDP which consists of state $s_t$, action $\boldsymbol{a}_t$, stationary transition dynamics distribution $p(s_{t+1}|s_t,\boldsymbol{a}_t)$ for any trajectory $(s_1,\boldsymbol{a}_1,\dots,s_t,\boldsymbol{a_t},\dots,s_{t^e},\boldsymbol{a}_{t^e})$, and reward function $r_t$, where $t\in[t^s,t^{s+\triangle t},\dots,t^{s+n\triangle t},\dots,t^{s+(N-1)\triangle t}]$. A stochastic policy $\pi_\theta(\boldsymbol{a}_t|s_t)$ is used to select actions in MDP. It is the set of probability measures on $\boldsymbol{a}_t$ associated with the policy when the agent is at state $s_t$ and determined by a set of parameters $\theta \in \mathbb{R}^n$. The parameters $\theta$ are implemented by the neural network and can be adjusted by agent to affect the policy.

\textit{2) Action-Value Function}

The agent follows the policy $\pi_\theta(\boldsymbol{a}_t|s_t)$ to interact with the environment, which is the demand bidding market and the consumers' interaction in our problem. The reward and a trajectory are gained sequentially. Cumulative discounted reward of this trajectory is defined as

\begin{equation}
    R_t=\sum_{k=t}^{\infty}\gamma^{k-t}r_k,
\end{equation}


where $\gamma \in [0,1]$ is a discounted factor for the agent to decide the weight of focusing on the future reward. The objective for the model is to maximize $R_t$. The action-value function,
\begin{equation}
\small 
 \begin{aligned}
    Q^\pi (s_t,a_t)\!&=\!E_{\boldsymbol{s}\sim\tau^\pi,\boldsymbol{a}\sim \pi_\theta}[R_t|s_t,\boldsymbol{a}_t]\\
    &=\!E_{\boldsymbol{s}\sim\tau^\pi,\boldsymbol{a}\sim \pi_\theta}[r(s_t,\boldsymbol{a}_t)+\gamma E_{\boldsymbol{s}\sim\tau^\pi,\boldsymbol{a}\sim \pi^\theta}[Q^\pi (s_{t+1},\boldsymbol{a}_{t+1})]],\!
\end{aligned}
\label{eq:action}
\end{equation}
stands for the expected discounted return while $\tau^\pi$ is denoted as the state visitation distribution after choosing action $a_t$ in state $s_t$ following the policy $\pi_\theta$. The function is utilized to evaluate the chosen action during the learning process and can also be expressed in a recursive relationship known as the Bellman equation. The agent learns the policy by maximizing Equation (\ref{eq:action}).


The DDPG algorithm we employed can be referred to \citep{lillicrap2015continuous}, which based on the structure of actor-critic algorithm along with deterministic action policy and policy gradient update. The advantages of off-policy training, e.g. replaying buffer and targeting Q network from DQN, are utilized. Actions  $\boldsymbol{a}_t$ are stochastically sampled from the distribution of policy $\pi_\theta(\boldsymbol{a}_t|s_t)$ in actor-critic as $\small \boldsymbol{a}\sim \pi_\theta$. With the functions of DDPG, e.g., adding noise to deterministic actor policy, replay buffer, and parameter updating, our continuous bidding problem can be efficiently solved and converged. Online learning in Fig. \ref{fig:online} can also be thus achieved on daily base.
\vspace{-0.5cm}

\section{Environment Simulator and Simulations}
\label{Sim}
To test the robustness and learning ability of our method, two simulators are developed and detailed before the simulation. The results of offline learning over learning cases are shown to demonstrate the learning ability. Moreover, we use the trained model to do online learning and test the effectiveness of the system in the meantime.

\subsection{Environment Simulator}
The market clearing price $\lambda_{M,t}$ and the actual curtailment quantity of aggregated customers $q_{act,t}$ are the required data when training our proposed model. As the robustness target of our proposed model, the realistic data based simulators are proposed. The price simulator is constructed by historical MCP data from the bidding platform of Taiwan Power Company, while the curtailment simulator imitates the real consumption of each participants according to \citep{song2017price}.

Specifically, we adopt a second order surface fitting function of $V_t$ to build the price simulator based on the historical data. The MCP $\lambda_{M,t}$ can be derived under arbitrary simulated cases with input of period $t$ and system reserve rate $V_t$ to verify whether the bidding price $\lambda_{bid,t}$ from the agent wins the bid. Given historical MCP data of demand bidding market and the corresponding reserve rate in 2017 from Taiwan Power Company, we adopt second order surface fitting function to construct the simulator for the MCP $\lambda_{M,t}$: 
\begin{equation}
\lambda_{M,t}= p_1 \times t^2+p_2 \times V_t^2 + p_3 \times V_t \times t+p_4 \times t+p_5 \times V_t+p_6.
\label{simu1}
\end{equation}
Least square method is used to derive the parameters $[p_1,p_2,\dots,p_6]$ of the function and the results are shown in TABLE \ref{tb:para}. By learning the data, the agent is educated to capture the inner relationship as (\ref{simu1}). To estimate the robustness of the learning, a noise is also randomly added following a normal distribution. The range of the noises is determined based on the difference between realistic data and the simulated data.

\begin{table}[]
\centering \caption{The Parameter Settings of the Simulators}
\begin{tabular}{|c|c|c|c|}
\hline
\multicolumn{4}{|c|}{Parameters of the Simulators}                                        \\ \hline
\multicolumn{2}{|c|}{MCP Simulator} & \multicolumn{2}{c|}{Curtailment Simulator}          \\ \hline
$p_1$               &  -0.00042              & $\epsilon_{i,P}$                         &  [-0.4,0]                         \\ \hline
$p_2$               &  126.7125              &  $\epsilon_{i,SP}$                          &  [-0.6,-0.4]                         \\ \hline
$p_3$               &    -0.06412            &    $\epsilon_{i,OP}$                     &    [-1,-0.6]                       \\ \hline
$p_4$               &  0.04937              &\multicolumn{2}{c|}{\cellcolor[HTML]{C0C0C0}}                           \\ \cline{1-2}
$p_5$               &   -55.07590              & \multicolumn{2}{c|}{\cellcolor[HTML]{C0C0C0}}                   \\ \cline{1-2}
$p_6$                &  7.45740             & \multicolumn{2}{c|}{\multirow{-2}{*}{\cellcolor[HTML]{C0C0C0}}} \\ \hline 
\end{tabular}
\label{tb:para}
\end{table}

Regarding customers’ actual consumption during DR event, the relationship between their electricity consumption quantity 
$p_{i,t}$ and their bidding price $\lambda_{i,t}$ can be modeled through the coefficients of price elasticity $\epsilon_{i,t}\in[-1,0]$ describing how sensitive their demands would change in response to their bidding price:
\begin{equation}
    \epsilon_{i,t}=\frac{(p_{i,t}-p^0_{i,t})/p^0_{i,t}}{\vert \lambda_{i,t}-\lambda_{ToU,t}\vert/\lambda_{ToU,t} },
\end{equation}
where $p^0_{i,t}$ indicates the original load for customer $i$ at time $t$ obtaining from the load forecasting and $\lambda_{ToU,t}$ refers the Time of Use rate of the consumption \citep{song2017price}. As estimated in Table \ref{tb:para}, the range of $\epsilon_{i,t}$ varies according to the time intervals, e.g., peak (P), off-peak(OP), and semi-peak(SP). The elasticity of changing the load during OP is highest. 
And there are 16 customers are assumed aggregated under the aggregator with individual $\epsilon_{i,t}$, which is constant during the time interval $t$ for each consumer $i$. 
\subsection{Offline Learning}
We apply grid search \citep{bergstra2011algorithms} on the validation dataset and $Success Rate \geq 90\% $ as the evaluation rule. $Success Rate$ is calculated by how much percentage of periods can earn profit among the total training periods. We found that at least 120 data are needed to train a qualified model. That is, 15 days dataset of DR events,  which contains a total of 120 periods (15 mins per data and 2 hours considered in a day) rom simulators, 
is needed to iteratively train the bidding system. As the noises found to range from $0-0.5$, three scenarios for validating the robustness are proposed with diverse noise modeled by normal distribution $N\sim(\mu,\sigma^2)$ to the simulated data, e.g., standard deviation $\sigma$ equals to 0, 0.2, and 0.5.

\subsubsection{Bidding Price and Quantity}
\begin{figure}[ht]
    \centering
    \includegraphics[width=1\columnwidth]{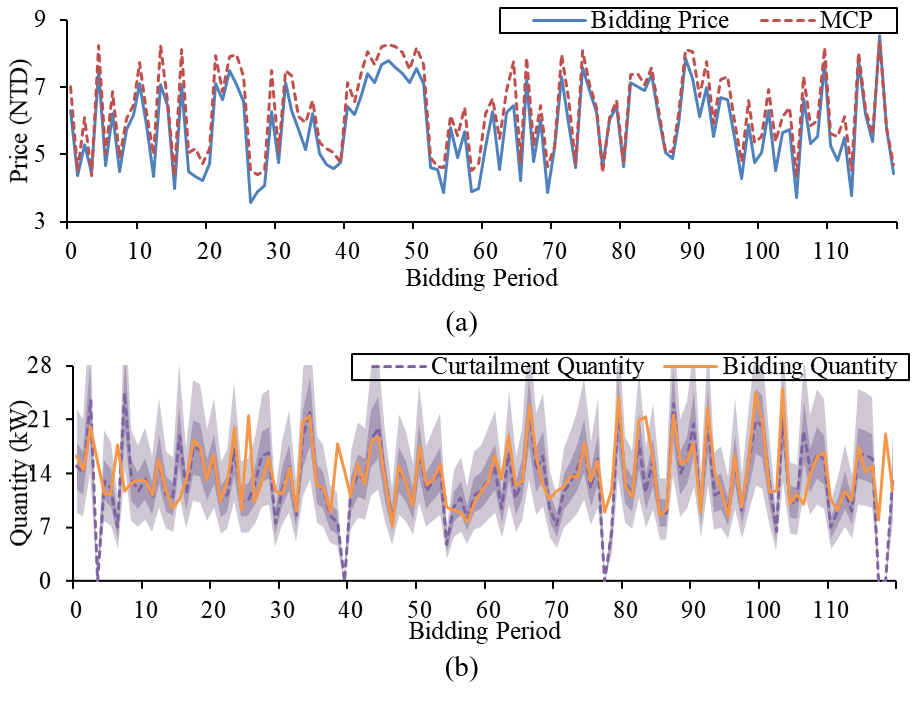}
    \caption{Offline learning results for Scenario 1}
    \label{fig:offline1}
\end{figure}

\begin{figure}[ht]
    \centering
    \includegraphics[width=1\columnwidth]{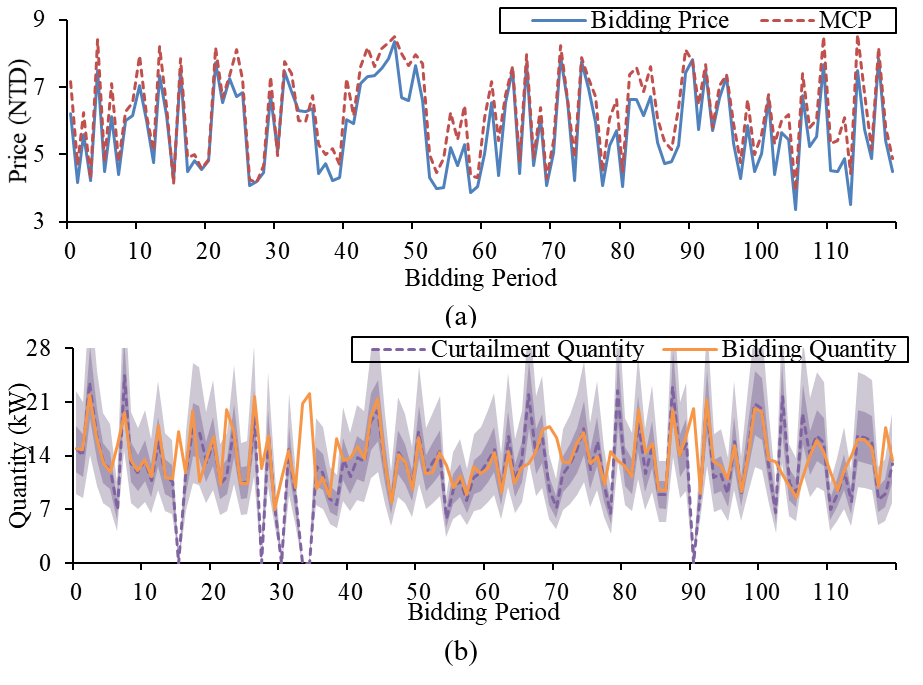}
    \caption{Offline learning results for Scenario 2}
    \label{fig:offline2}
\end{figure}

\begin{figure}[ht]
    \centering
    \includegraphics[width=1\columnwidth]{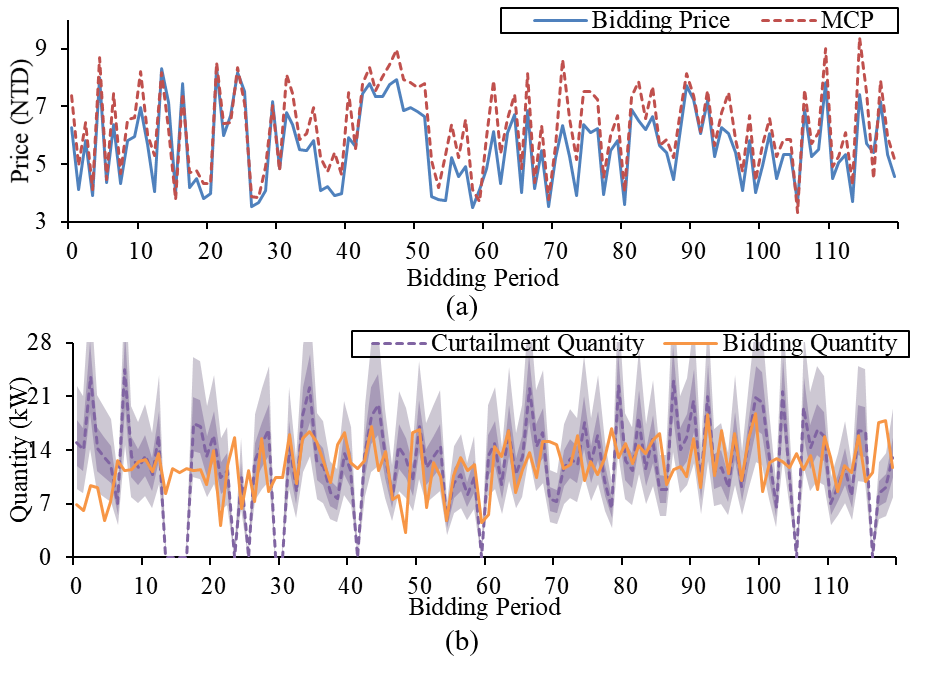}
    \caption{Offline learning results for Scenario 3}
    \label{fig:offline3}
\end{figure}

After offline training, the testing results of bidding price and bidding quantity for three scenarios in 120 time intervals are shown in Fig. \ref{fig:offline1}, \ref{fig:offline2}, and \ref{fig:offline3} respectively. Take the results in Fig. \ref{fig:offline1}(a) as the example, the bidding price made by the bidding system in the blue line can be maintained slightly lower than the MCP in the red dash line, namely successfully tracing the MCP when confronting the state $s_t$ from the environment. The bidding quantity made by the bidding system represented as the orange line is nearly equal to the actual curtailment represented as the purple dash line in Fig. \ref{fig:offline1}(b). Moreover, according to the incentive ratio $\alpha_t$ demonstrated in Equation \ref{intratio} and involved in the reward function, the range between 0.8 and 1.2 of the actual curtailment is represented by the lavender area while the range between 0.6 and 1.4 is displayed by lilac area.  It demonstrates the ability of agent structure employed to predict the actual curtailment quantity of aggregated customers. Particularly, though some bidding quantities are not exactly the same as the actual curtailment, they lie in between the range of 0.8 and 1.2 in most of the situations.


A similar performance is revealed in Fig. \ref{fig:offline2}, and \ref{fig:offline3}.  The proposed method employed does capture all the trends and relationships within the reserve rate, time stamp, and offered price by the customers. Ultimately, even the noises, e.g., $\sigma=0.2$ and $0.5$, added to the simulator, our proposed method achieves good performance and performs outstanding robustness. 

Overall, the details of offline learning results for three scenarios are shown in TABLE \ref{tb:comparison}. The success rate of Scenario 2 is 97.5\% which is quite near to 98.3\% of Scenario 1. Even for Scenario 3, the case with the highest noise in the simulator, the success rate still can be maintained as 93.3\%. All effectively demonstrate the robustness of learning ability for the proposed bidding system. For the execution rate $\xi$, it demonstrates the performance of the bidding quantity. The aggregator can gain a bonus when $0.6\leq\xi\leq1.5$ referring to the DR rule shown in TABLE I. Thus, we calculate the achievement rate of $0.8\leq\xi\leq1.2$ and $0.6\leq\xi\leq1.5$. The results turn out that the average rate for the former one is around 73\%, and the latter one reaches around 82\% for the three scenarios, which shows good predictivity of the method employed under a limited amount of 120 periods of data.

\begin{table}[t]
\centering
\caption{Offline learning results for three scenarios}
\setlength{\tabcolsep}{3.2mm}{
\begin{tabular}{cccc}
\hline

\rowcolor[HTML]{C0C0C0} 
\multicolumn{1}{c}{\cellcolor[HTML]{C0C0C0}\textbf{Scenario}} & \multicolumn{1}{c}{\cellcolor[HTML]{C0C0C0}\textbf{Success rate}} & \multicolumn{1}{c}{\cellcolor[HTML]{C0C0C0}\textbf{\begin{tabular}[c]{@{}c@{}}Rate of\\ $0.8\leq\xi\leq1.2$\end{tabular}}} & \multicolumn{1}{c}{\cellcolor[HTML]{C0C0C0}\textbf{\begin{tabular}[c]{@{}c@{}}Rate of\\ $0.6\leq\xi\leq1.5$\end{tabular}}} \\ \hline\hline
\rowcolor[HTML]{EFEFEF} 1& 98.3\%& 85.4\%& 90.0\% \\ \hline
\rowcolor[HTML]{FFFFFF} 
2& 97.5\%& 83.3\%& 85.5\%\\ \hline
\rowcolor[HTML]{EFEFEF} 
3& 93.3\%& 51.1\%& 70.6\%\\ \hline
\end{tabular}}
\label{tb:comparison}
\end{table}

\subsubsection{Cumulative Profit over Iteration}
\label{sec:cum}
\begin{figure}[ht]
    \centering
    \includegraphics[width=1\columnwidth]{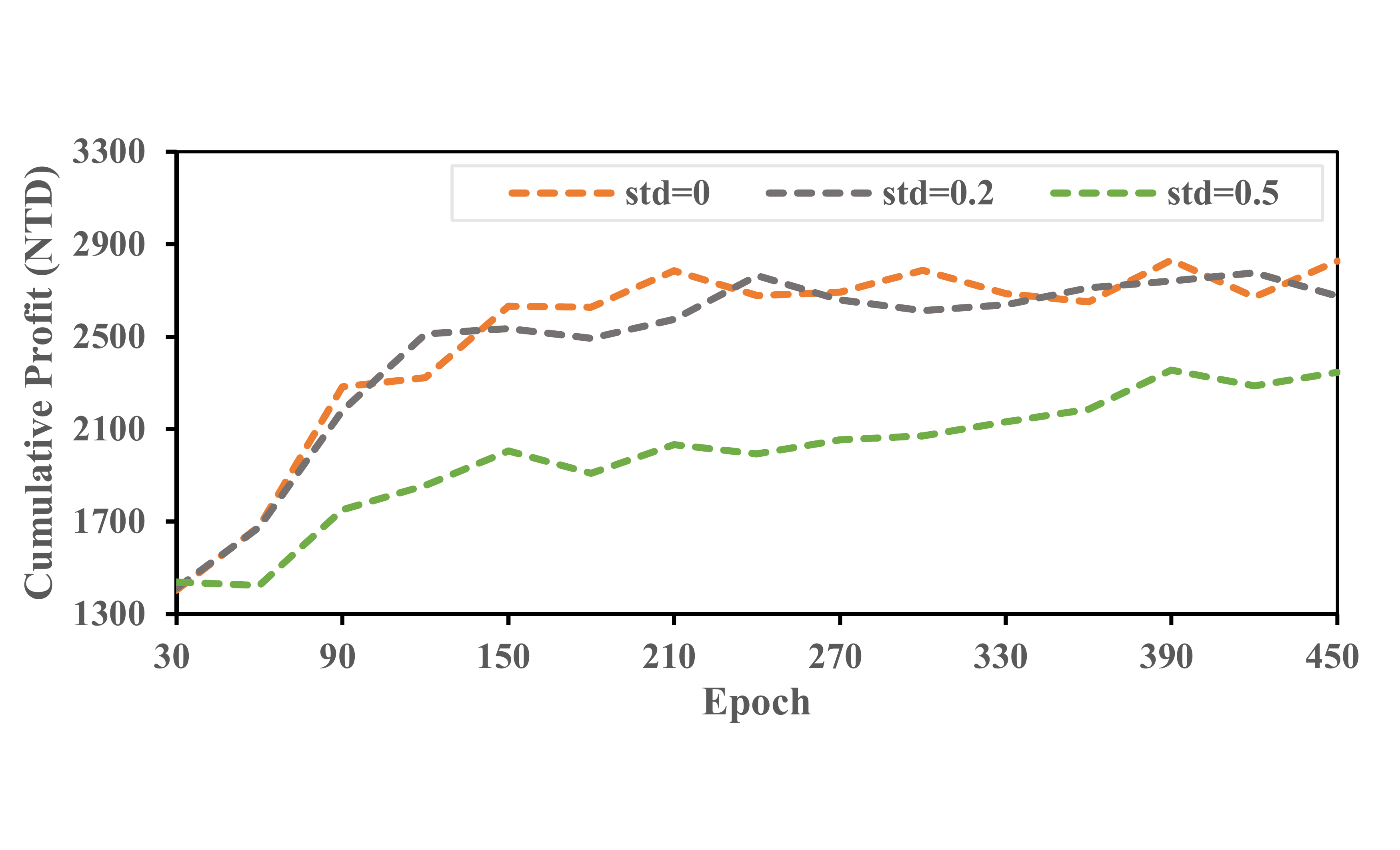}
    \caption{Cumulative profit over the offline learning eposide}
    \label{fig:offacc}
\end{figure}

Cumulative profit is defined as the total profit the aggregator can earn for each episode by the proposed learning structure. The cumulative profits, changing by the training for three scenarios are shown in Fig. \ref{fig:offacc} with the orange line for Scenario 1, the grey line for Scenario 2, and the green line for Scenario 3.
We can see that the cumulative profit increases with the training processing. It gradually becomes stable within a range after 120-th episode which reveals the bidding system has learned to determine a bidding strategy that is approach to optimal solution. 


\subsection{Online Learning Results}
\begin{figure}[ht]
    \centering
    \includegraphics[width=1\columnwidth]{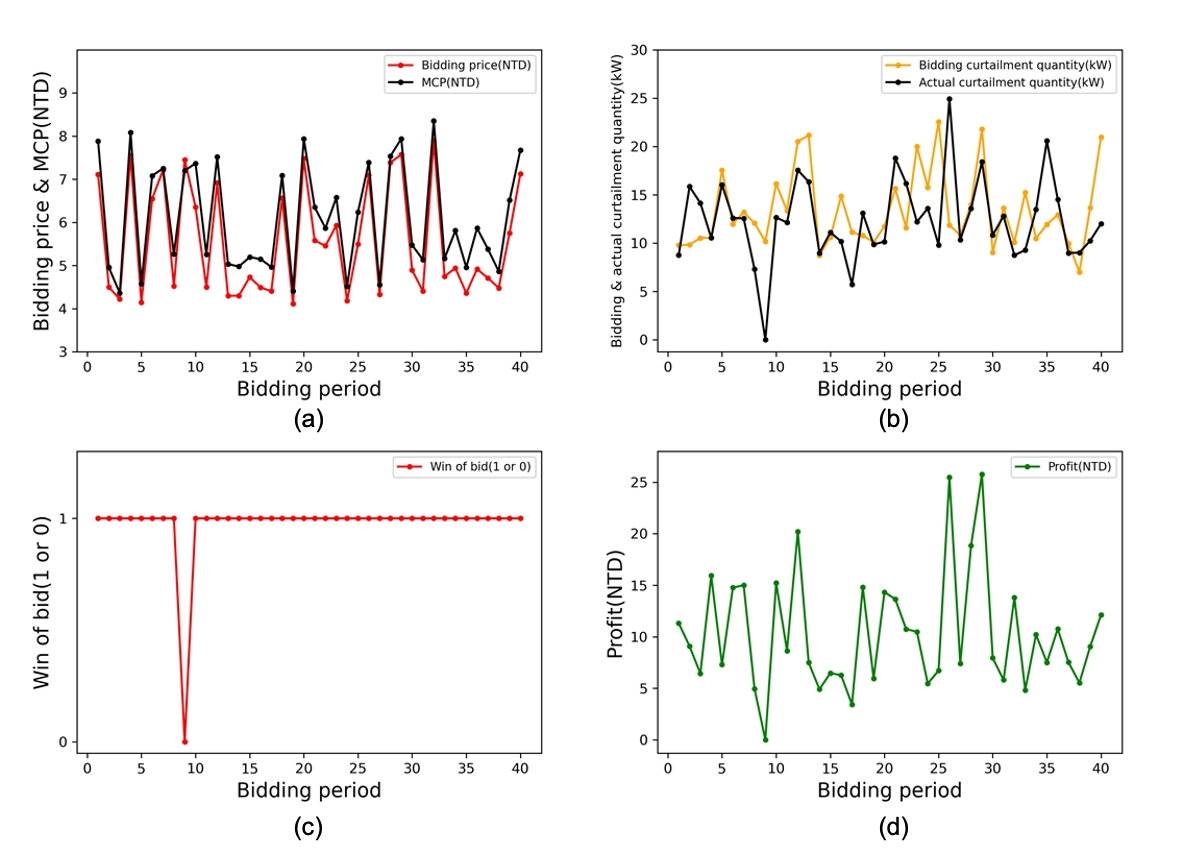}
    \caption{Online learning results for Scenario 1}
    \label{fig:online1}
\end{figure}

\begin{figure}[ht]
    \centering
    \includegraphics[width=1\columnwidth]{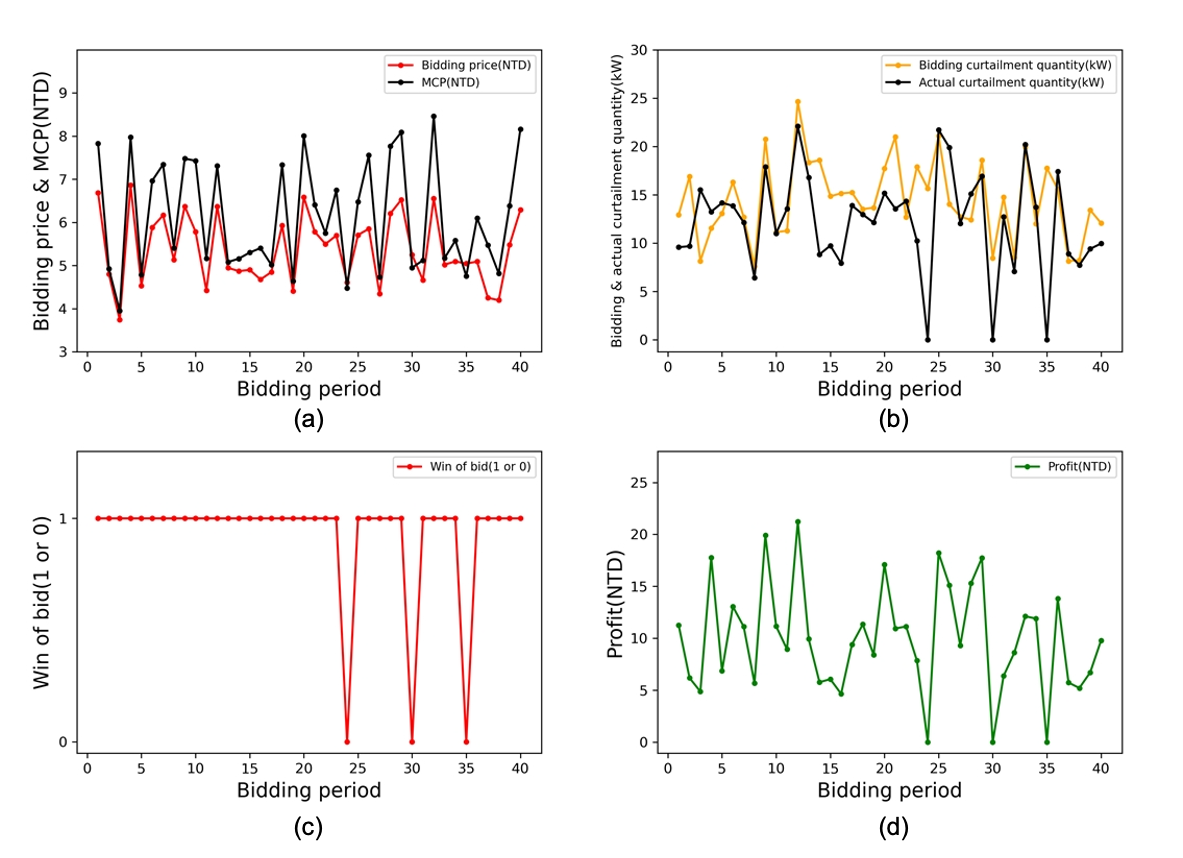}
    \caption{Online learning results for Scenario 2}
    \label{fig:online2}
\end{figure}

\begin{figure}[ht]
    \centering
    \includegraphics[width=1\columnwidth]{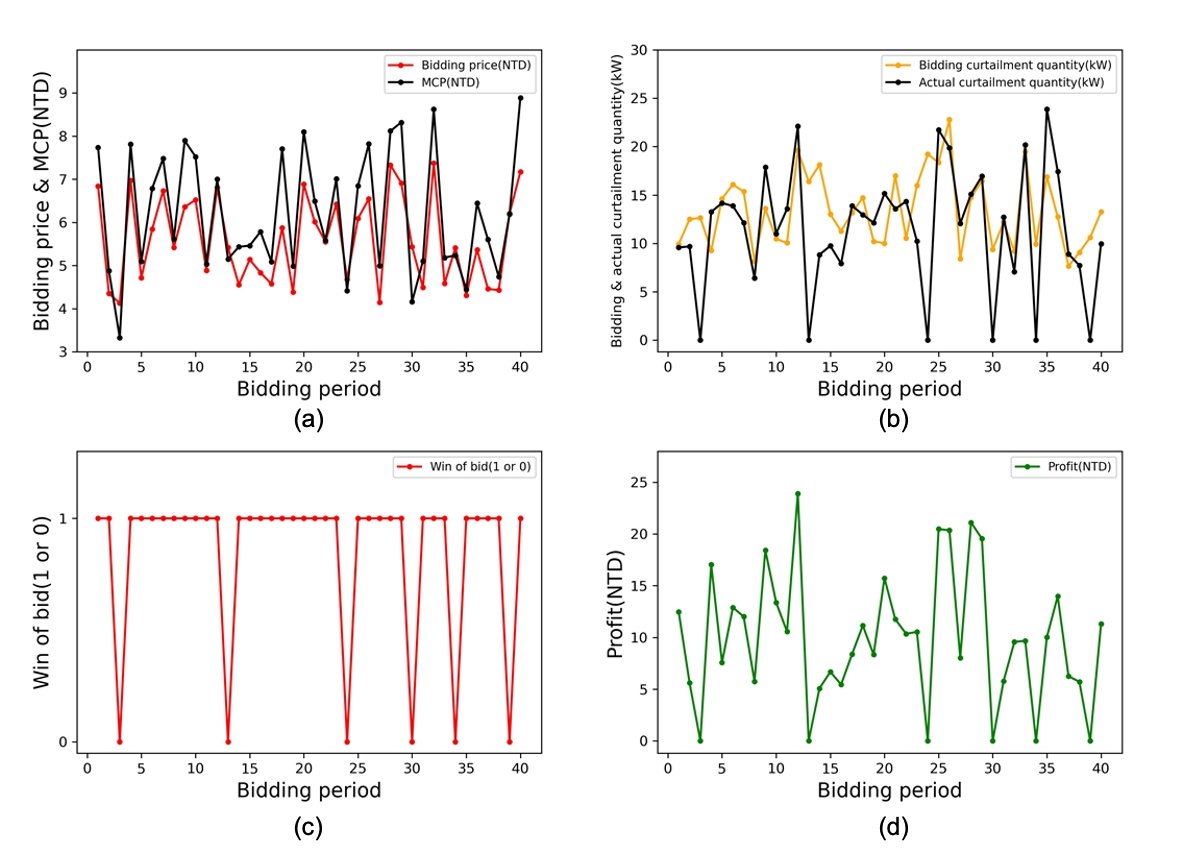}
    \caption{Online learning results for Scenario 3}
    \label{fig:online3}
\end{figure}

With the same evaluation mechanism as offline learning while $Success Rate \geq 85\% $ is set, 50 days dataset for offline training and 5 days dataset for online training are found to be required for a qualified online bidding system. The parameters of the model are firstly obtained by the offline and sequential developed by online training and then tested in the online mode. Now, we test the model for 5 days which contains a total of 40 periods and  verify the performance of online learning. Online learning is feasible to the real situation that the aggregator participates in the bidding market. In addition to the results shown in Section \ref{sec:cum} have fully demonstrated the robustness and adaptivity of the proposed bidding system from 3 scenarios, we consider the 3 scenarios with the same standard deviation setting in this section to test the efficiency of online learning. The performance is shown in Fig. \ref{fig:online1}, \ref{fig:online2}, and \ref{fig:offline3}. As estimated (a) and (b) in the figures, not only the bidding price but the quantity, the strategy made by our model can markedly capture the trend and accordingly submit bidding. Specifically, the performance of the bidding strategy can be evaluated from the approaching degree from bidding price to the MCP and the difference between actual curtailment and bidding quantity. Moreover, we evaluate the situation where the aggregator wins the bid and customers have the actual curtailment at time $t$ as a win deal. The win flag in (c) of Fig. \ref{fig:online1}, \ref{fig:online2}, and \ref{fig:online3} estimates a win deal if the value is 1 and the corresponding profit for aggregator is shown in (d). It
represents high bidding winning rates, e.g., 97.5\%, 92.5\%, and 85.0\%, which is to say only one failure during 40 bidding periods in Scenario 1. Nevertheless, there is no significant gap between the profits owned by the bidding strategy in 3 scenarios from the observation in (d) of the three figures. 

The proposed model is compared with a neural network (NN) based benchmark model. Two NNs, namely price model and quantity model, are employed to determine the bidding price and bidding quantity individually. 
The hyperparameters are set the same as the network in our model, e.g., 4 layers and the layers separately have 300, 600, 400, and 200 neurons. In other words, compared with our model the networks do not share the input or loss function but make the decision distinctly, which is the general structure to solve these kinds of problems. The training data for the models are the same as the proposed bidding model where the factors considered are time $t$, weekday/weekend category $w$, and reserve rate $V_t$ for the price model while $t$, $w$, and report prices from customers $\lambda_{i,t}$ for the quantity model. We list the average results of the online learning results for our model and benchmark model in 3 scenarios in Table \ref{tb:comparison2}. 
The success rate over 40 periods obtained by proposed is up to 97.50\% shown in the table, which ensuring the aggregator can nearly earn a profit every time and learn the trend of MCP. And the execution rate is 76.9\% in the range $0.8\leq\xi\leq1.2$, which advances the profit increase. On the other hand, the performance of the benchmark model, which without the experience sharing among the agents nor the interaction considered in the situation, only achieves 42.5\% of success rate and 57.5\% of execution rate in the range of $0.8\leq\xi\leq1.2$. Overall, the proposed two-agent DDPG scheme outperforms superiorly to the NN-based model.

\begin{table}[h!]
\centering
\caption{The Average Performance Comparison between The Proposed Method and Benchmark}
\setlength{\tabcolsep}{7.5mm}{
\begin{tabular}{ccc}
\hline

\rowcolor[HTML]{C0C0C0} 
\multicolumn{1}{c}{\cellcolor[HTML]{C0C0C0}\textbf{Method}} & \multicolumn{1}{c}{\cellcolor[HTML]{C0C0C0}\textbf{Success rate}} & \multicolumn{1}{c}{\cellcolor[HTML]{C0C0C0}\textbf{\begin{tabular}[c]{@{}c@{}}Rate of\\ $0.8\leq\xi\leq1.2$\end{tabular}}} \\ \hline\hline
\rowcolor[HTML]{EFEFEF} Proposed& 97.5\%& 76.9\% \\ \hline
\rowcolor[HTML]{FFFFFF} 
Benchmark & 42.5\%& 57.5\%\\ \hline
\end{tabular}}
\label{tb:comparison2}
\end{table}

\section{Conclusion}
\label{con}
We have developed a machine learning-based bidding and purchasing methodology for aggregators to participate in the DR program. The state-of-the-art DDPG reinforcement learning algorithm has been utilized and developed into a two-agent structure. The model refines and updates the bidding decisions through offline and online learning with the bidding experience. The uncertainties from the MCP and the customers’ consumption behavior confronted by the aggregator have thus been well measured and considered in the reward functions. 
Besides, to simulate the real situations the aggregator may face and to test the robustness of our proposed scheme, the simulators of MCP and customers' consumption reduction are introduced. Among all the scenarios, the proposed method has been proved to achieve a good performance and have a stable bidding ability especially after more and more bidding experiences are involved. In particular, the indices of bidding success rate and execution rate achieved from the proposed method perform well, as observed from the scenarios even with higher uncertainties. The results further verify the capability and robustness of the proposed method.

For future work, based on the work we already developed, we plan to apply our online interactive model to other types of DR programs, e.g., price-driven and incentive-driven programs. As estimated, our model achieved good performance on learning the consumption behaviors behind the bidding plan submitted by the consumer. However, our study is based on the rule that the consumer needs to submit the willing price to the aggregator.  It makes some limitations on the application. By utilizing the model and learning the reaction of consumers to the price or incentive signal, we believe the DR quantity and potential of end-users developed in other DR programs can be also well evaluated. 
\vspace{-0.5cm}




\bibliographystyle{IEEEtran} 
 \bibliography{ref.bib}

\begin{thebibliography}{10}
\providecommand{\url}[1]{#1}
\csname url@samestyle\endcsname
\providecommand{\newblock}{\relax}
\providecommand{\bibinfo}[2]{#2}
\providecommand{\BIBentrySTDinterwordspacing}{\spaceskip=0pt\relax}
\providecommand{\BIBentryALTinterwordstretchfactor}{4}
\providecommand{\BIBentryALTinterwordspacing}{\spaceskip=\fontdimen2\font plus
\BIBentryALTinterwordstretchfactor\fontdimen3\font minus
  \fontdimen4\font\relax}
\providecommand{\BIBforeignlanguage}[2]{{%
\expandafter\ifx\csname l@#1\endcsname\relax
\typeout{** WARNING: IEEEtran.bst: No hyphenation pattern has been}%
\typeout{** loaded for the language `#1'. Using the pattern for}%
\typeout{** the default language instead.}%
\else
\language=\csname l@#1\endcsname
\fi
#2}}
\providecommand{\BIBdecl}{\relax}
\BIBdecl

\bibitem{lu2018evaluation}
S.~D. Lu, M.~T. Kuo, C.~C. Wu, M.~C. Tsou, and H.~Zeng, ``Evaluation of the
  maximum allowable generation capacity after offshore wind farms are
  integrated into the existing taipower system-a case study in taiwan,'' in
  \emph{2018 IEEE/IAS 54th Industrial and Commercial Power Systems Technical
  Conference (I\&CPS)}.\hskip 1em plus 0.5em minus 0.4em\relax IEEE, 2018, pp.
  1--15.

\bibitem{strasser2014review}
T.~Strasser, F.~Andr{\'e}n, J.~Kathan, C.~Cecati, C.~Buccella, P.~Siano,
  P.~Leitao, G.~Zhabelova, V.~Vyatkin, P.~Vrba \emph{et~al.}, ``A review of
  architectures and concepts for intelligence in future electric energy
  systems,'' \emph{IEEE Transactions on Industrial Electronics}, vol.~62,
  no.~4, pp. 2424--2438, 2014.

\bibitem{8681637}
M.~Shafie-khah, P.~Siano, J.~Aghaei, M.~A.~S. Masoum, F.~Li, and J.~P.~S.
  Catalão, ``Comprehensive review of the recent advances in industrial and
  commercial dr,'' \emph{IEEE Transactions on Industrial Informatics}, vol.~15,
  no.~7, pp. 3757--3771, 2019.

\bibitem{iria2018optimal}
J.~Iria, F.~Soares, and M.~Matos, ``Optimal supply and demand bidding strategy
  for an aggregator of small prosumers,'' \emph{Applied Energy}, vol. 213, pp.
  658--669, 2018.

\bibitem{utama2021demand}
C.~Utama, S.~Troitzsch, and J.~Thakur, ``Demand-side flexibility and
  demand-side bidding for flexible loads in air-conditioned buildings,''
  \emph{Applied Energy}, vol. 285, p. 116418, 2021.

\bibitem{dong2021strategic}
Y.~Dong, Z.~Dong, T.~Zhao, and Z.~Ding, ``A strategic day-ahead bidding
  strategy and operation for battery energy storage system by reinforcement
  learning,'' \emph{Electric Power Systems Research}, vol. 196, p. 107229,
  2021.

\bibitem{EIA2018}
\BIBentryALTinterwordspacing
U.~E.~I. Administration. (2018) How much energy is consumed in u.s. residential
  and commercial buildings? [Online]. Available:
  \url{https://www.eia.gov/tools/faqs/faq.php?}
\BIBentrySTDinterwordspacing

\bibitem{bahrami2017online}
S.~Bahrami, V.~W. Wong, and J.~Huang, ``An online learning algorithm for demand
  response in smart grid,'' \emph{IEEE Transactions on Smart Grid}, vol.~9,
  no.~5, pp. 4712--4725, 2017.

\bibitem{parizy2018low}
E.~S. Parizy, H.~R. Bahrami, and S.~Choi, ``A low complexity and secure demand
  response technique for peak load reduction,'' \emph{IEEE Transactions on
  Smart Grid}, vol.~10, no.~3, pp. 3259--3268, 2018.

\bibitem{lu2012evaluation}
N.~Lu, ``An evaluation of the hvac load potential for providing load balancing
  service,'' \emph{IEEE Transactions on Smart Grid}, vol.~3, no.~3, pp.
  1263--1270, 2012.

\bibitem{7579628}
S.~{Moon} and J.~{Lee}, ``Multi-residential demand response scheduling with
  multi-class appliances in smart grid,'' \emph{IEEE Transactions on Smart
  Grid}, vol.~9, no.~4, pp. 2518--2528, 2018.

\bibitem{korkas2016occupancy}
C.~D. Korkas, S.~Baldi, I.~Michailidis, and E.~B. Kosmatopoulos,
  ``Occupancy-based demand response and thermal comfort optimization in
  microgrids with renewable energy sources and energy storage,'' \emph{Applied
  Energy}, vol. 163, pp. 93--104, 2016.

\bibitem{lyao2018}
L.~{Yao} and W.~H. {Lim}, ``Optimal purchase strategy for demand bidding,''
  \emph{IEEE Transactions on Power Systems}, vol.~33, no.~3, pp. 2754--2762,
  2018.

\bibitem{jia2019purchase}
Y.~Jia, Z.~Mi, Y.~Yu, Z.~Song, L.~Liu, and C.~Sun, ``Purchase bidding strategy
  for load agent with the incentive-based demand response,'' \emph{IEEE
  Access}, vol.~7, pp. 58\,626--58\,637, 2019.

\bibitem{li2017distributed}
P.~Li, H.~Wang, and B.~Zhang, ``A distributed online pricing strategy for
  demand response programs,'' \emph{IEEE Transactions on Smart Grid}, vol.~10,
  no.~1, pp. 350--360, 2019.

\bibitem{tang2019optimal}
W.~Tang and H.-T. Yang, ``Optimal operation and bidding strategy of a virtual
  power plant integrated with energy storage systems and elasticity demand
  response,'' \emph{IEEE Access}, vol.~7, pp. 79\,798--79\,809, 2019.

\bibitem{fazlalipour2019risk}
P.~Fazlalipour, M.~Ehsan, and B.~Mohammadi-Ivatloo, ``Risk-aware stochastic
  bidding strategy of renewable micro-grids in day-ahead and real-time
  markets,'' \emph{Energy}, vol. 171, pp. 689--700, 2019.

\bibitem{zhao2018strategic}
T.~Zhao, X.~Pan, S.~Yao, C.~Ju, and L.~Li, ``Strategic bidding of hybrid ac/dc
  microgrid embedded energy hubs: A two-stage chance constrained stochastic
  programming approach,'' \emph{IEEE Transactions on Sustainable Energy},
  vol.~11, no.~1, pp. 116--125, 2018.

\bibitem{alashery2019second}
M.~K. AlAshery, D.~Xiao, and W.~Qiao, ``Second-order stochastic dominance
  constraints for risk management of a wind power producer's optimal bidding
  strategy,'' \emph{IEEE Transactions on Sustainable Energy}, 2019.

\bibitem{7216732}
M.~{Wei} and J.~{Zhong}, ``Optimal bidding strategy for demand response
  aggregator in day-ahead markets via stochastic programming and robust
  optimization,'' in \emph{2015 12th International Conference on the European
  Energy Market (EEM)}, 2015, pp. 1--5.

\bibitem{liu2015bidding}
G.~Liu, Y.~Xu, and K.~Tomsovic, ``Bidding strategy for microgrid in day-ahead
  market based on hybrid stochastic/robust optimization,'' \emph{IEEE
  Transactions on Smart Grid}, vol.~7, no.~1, pp. 227--237, 2015.

\bibitem{rahimiyan2015strategic}
M.~Rahimiyan and L.~Baringo, ``Strategic bidding for a virtual power plant in
  the day-ahead and real-time markets: A price-taker robust optimization
  approach,'' \emph{IEEE Transactions on Power Systems}, vol.~31, no.~4, pp.
  2676--2687, 2015.

\bibitem{asensio2015risk}
M.~Asensio and J.~Contreras, ``Risk-constrained optimal bidding strategy for
  pairing of wind and demand response resources,'' \emph{IEEE Transactions on
  Smart Grid}, vol.~8, no.~1, pp. 200--208, 2015.

\bibitem{xu2015risk}
Z.~Xu, Z.~Hu, Y.~Song, and J.~Wang, ``Risk-averse optimal bidding strategy for
  demand-side resource aggregators in day-ahead electricity markets under
  uncertainty,'' \emph{IEEE Transactions on Smart Grid}, vol.~8, no.~1, pp.
  96--105, 2015.

\bibitem{nguyen2014risk}
D.~T. Nguyen and L.~B. Le, ``Risk-constrained profit maximization for microgrid
  aggregators with demand response,'' \emph{IEEE Transactions on smart grid},
  vol.~6, no.~1, pp. 135--146, 2014.

\bibitem{henderson2018deep}
P.~Henderson, R.~Islam, P.~Bachman, J.~Pineau, D.~Precup, and D.~Meger, ``Deep
  reinforcement learning that matters,'' in \emph{Thirty-Second AAAI Conference
  on Artificial Intelligence}, 2018.

\bibitem{vazquez2019reinforcement}
J.~R. V{\'a}zquez-Canteli and Z.~Nagy, ``Reinforcement learning for demand
  response: A review of algorithms and modeling techniques,'' \emph{Applied
  energy}, vol. 235, pp. 1072--1089, 2019.

\bibitem{wang2020deep}
B.~Wang, Y.~Li, W.~Ming, and S.~Wang, ``Deep reinforcement learning method for
  demand response management of interruptible load,'' \emph{IEEE Transactions
  on Smart Grid}, vol.~11, no.~4, pp. 3146--3155, 2020.

\bibitem{lu2019demand}
R.~Lu, S.~H. Hong, and M.~Yu, ``Demand response for home energy management
  using reinforcement learning and artificial neural network,'' \emph{IEEE
  Transactions on Smart Grid}, vol.~10, no.~6, pp. 6629--6639, 2019.

\bibitem{lu2019incentive}
R.~Lu and S.~H. Hong, ``Incentive-based demand response for smart grid with
  reinforcement learning and deep neural network,'' \emph{Applied energy}, vol.
  236, pp. 937--949, 2019.

\bibitem{rahimiyan2008supplier}
M.~Rahimiyan and H.~R. Mashhadi, ``Supplier's optimal bidding strategy in
  electricity pay-as-bid auction: Comparison of the q-learning and a
  model-based approach,'' \emph{Electric Power Systems Research}, vol.~78,
  no.~1, pp. 165--175, 2008.

\bibitem{yan2018review}
X.~Yan, Y.~Ozturk, Z.~Hu, and Y.~Song, ``A review on price-driven residential
  demand response,'' \emph{Renewable and Sustainable Energy Reviews}, vol.~96,
  pp. 411--419, 2018.

\bibitem{website2013}
\BIBentryALTinterwordspacing
Taipower company, demand-side management measures. [Online]. Available:
  \url{https://dbp.taipower.com.tw/TaiPowerDBP/Portal/Login}
\BIBentrySTDinterwordspacing

\bibitem{song2017price}
M.~Song and M.~Amelin, ``Price-maker bidding in day-ahead electricity market
  for a retailer with flexible demands,'' \emph{IEEE Transactions on power
  systems}, vol.~33, no.~2, pp. 1948--1958, 2018.

\bibitem{li2011modeling}
G.~Li, J.~Shi, and X.~Qu, ``Modeling methods for genco bidding strategy
  optimization in the liberalized electricity spot market--a state-of-the-art
  review,'' \emph{Energy}, vol.~36, no.~8, pp. 4686--4700, 2011.

\bibitem{caiso}
\BIBentryALTinterwordspacing
CAISO. California iso current and forecasted demand. [Online]. Available:
  \url{http://www.caiso.com/TodaysOutlook/Pages/default.aspx}
\BIBentrySTDinterwordspacing

\bibitem{lillicrap2015continuous}
T.~P. Lillicrap, J.~J. Hunt, A.~Pritzel, N.~Heess, T.~Erez, Y.~Tassa,
  D.~Silver, and D.~Wierstra, ``Continuous control with deep reinforcement
  learning,'' \emph{arXiv preprint arXiv:1509.02971}, 2015.

\bibitem{bergstra2011algorithms}
J.~S. Bergstra, R.~Bardenet, Y.~Bengio, and B.~K{\'e}gl, ``Algorithms for
  hyper-parameter optimization,'' in \emph{Advances in neural information
  processing systems}, 2011, pp. 2546--2554.

\end{thebibliography}






\end{document}